\documentclass{article}

\usepackage[preprint]{neurips_2025}
\makeatletter
\renewcommand{\@noticestring}{*\texttt{luo00318@umn.edu}.}
\makeatother

\usepackage[utf8]{inputenc} 
\usepackage[T1]{fontenc}    
\usepackage{hyperref}       
\usepackage{url}            
\usepackage{booktabs}       
\usepackage{amsfonts}       
\usepackage{nicefrac}       
\usepackage{microtype}      
\usepackage{xcolor}         
\usepackage{multirow}
\usepackage{multicol}
\usepackage{graphicx}
\usepackage{array}
\usepackage{caption}
\usepackage{xcolor}
\usepackage{fvextra}

\definecolor{promptbg}{HTML}{F7F7F9}
\definecolor{promptframe}{HTML}{D0D7DE}

\DefineVerbatimEnvironment{PromptVerbatim}{Verbatim}{
  breaklines=true,
  breakanywhere=true,
  fontsize=\footnotesize,
  frame=single,
  framesep=2mm,
  rulecolor=\color{promptframe},
  bgcolor=promptbg,
  bgcolorpadding=1mm
}
\setcitestyle{numbers,square,comma,sort&compress}

\title{\textbf{AgentDS Technical Report:\\ Benchmarking the Future of Human-AI Collaboration in Domain-Specific Data Science}}

\author{
\makebox[\textwidth][c]{
\begin{tabular}{>{\mdseries}c}
\bfseries
An Luo$^{1,*}$, Jin Du$^{1}$, Xun Xian$^{2}$, Robert Specht$^{1}$, 
Fangqiao Tian$^{1}$, Ganghua Wang$^{3}$, \\\bfseries 
Xuan Bi$^{4}$, Charles Fleming$^{5}$, Ashish Kundu$^{5}$,
Jayanth Srinivasa$^{5}$, \\\bfseries 
Mingyi Hong$^{6}$, Rui Zhang$^{7}$,
Tianxi Li$^{1}$, Galin Jones$^{1}$, Jie Ding$^{1,2}$\\[1ex]
$^{1}$School of Statistics, University of Minnesota \\
$^{2}$AIScientists, Inc. \\
$^{3}$Data Science Institute, University of Chicago \\
$^{4}$Carlson School of Management, University of Minnesota \\
$^{5}$Cisco Research \\
$^{6}$Department of Electrical and Computer Engineering, University of Minnesota \\
$^{7}$Division of Computational Health Sciences, University of Minnesota
\end{tabular}
}
}

\begin{document}
	
	\maketitle
	
\begin{abstract}

		Data science plays a critical role in transforming complex data into actionable insights across numerous domains. Recent developments in large language models (LLMs) and artificial intelligence (AI) agents have significantly automated data science workflow. However, it remains unclear to what extent AI agents can match the performance of human experts on domain-specific data science tasks, and in which aspects human expertise continues to provide advantages. We introduce \textbf{AgentDS}, a benchmark and competition designed to evaluate both AI agents and human-AI collaboration performance in domain-specific data science. AgentDS consists of \textbf{17 challenges across six industries}: commerce, food production, healthcare, insurance, manufacturing, and retail banking. We conducted an open competition involving \textbf{29 teams and 80 participants}, enabling systematic comparison between human-AI collaborative approaches and AI-only baselines. Our results show that current AI agents struggle with domain-specific reasoning. 
AI-only baselines perform below the top quartile of competition participants, while the strongest solutions arise from human-AI collaboration. 
        These findings challenge the narrative of complete automation by AI and underscore the enduring importance of human expertise in data science, while illuminating directions for the next generation of AI. Visit the AgentDS website \href{https://agentds.org/}{\textcolor{blue}{here}} and open source datasets \href{https://huggingface.co/datasets/lainmn/AgentDS}{\textcolor{blue}{here}}.
\end{abstract}
	
	\section{Introduction}
	
 Data science has become central to decision-making across industries, from healthcare diagnostics to financial risk assessment, where it blends statistics, computer science, and domain expertise to transform raw data into actionable insights ~\citep{Cao2017DataS,Grossi2021DataSA,Blair2019DataSO}. Recent advancement of large language models (LLMs) and AI agents demonstrate impressive capabilities in automating code generation and executing regular machine learning tasks ~\citep{achiam2023gpt,Anthropic2025claudecode,Hong2024DataIA,Li2024AutoKaggleAM,Jiang2025AIDEAE,Liang2025IMCTSEA,Grosnit2024LargeLM}. Some systems have even achieved Kaggle Grandmaster performance through structured reasoning ~\citep{Grosnit2024LargeLM}, while others automate data science workflows ~\citep{Seo2025SPIOEA,guo2024dsagent,Chi2024SELATE}. These advances suggest that many routine components of data science workflows may increasingly be automated, reducing the manual burden on human data scientists. 
 
Despite these advances in LLMs and AI agents for data science, a fundamental question remains unanswered: \textit{To what extent do human experts outperform autonomous AI agents on domain-specific data science tasks, and in which aspects does this advantage arise?} In practice, human data scientists consistently rely on specialized knowledge about data and tasks, incorporating crucial domain-specific nuances that enhance model performance ~\citep{Mao2019HowDS,Zhang2020HowDD,Lin2025spike,Lin2025spatial,Luo2025AssistedDSBH}. Such domain-driven decisions are often subtle yet essential, addressing complexities not captured by generic analytics workflows. However, current research on AI for data science has largely focused on generating generic code and pipeline executions ~\citep{Li2024AutoKaggleAM,Jiang2025AIDEAE}, often neglecting the domain-specific knowledge needed for real-world problems.
	
	Existing benchmarks for AI agents, while valuable, often do not test whether agentic AI can effectively leverage domain insights outside tabular data ~\citep{jing2025dsbench,chan2025mlebench,Hu2024InfiAgentDABenchEA,Zhang2025DataSciBenchAL,huang-etal-2024-da,pricope2025hardml}. Some recent work has demonstrated that current agentic AI typically generates generic code and pipeline executions, often neglecting the domain-specific knowledge needed for complex real-world problems ~\citep{Li2024AutoKaggleAM,Luo2025AssistedDSBH, luo2025agentds}. 
    
\begin{figure}[htb]
    \centering
\includegraphics[width=0.9\textwidth]{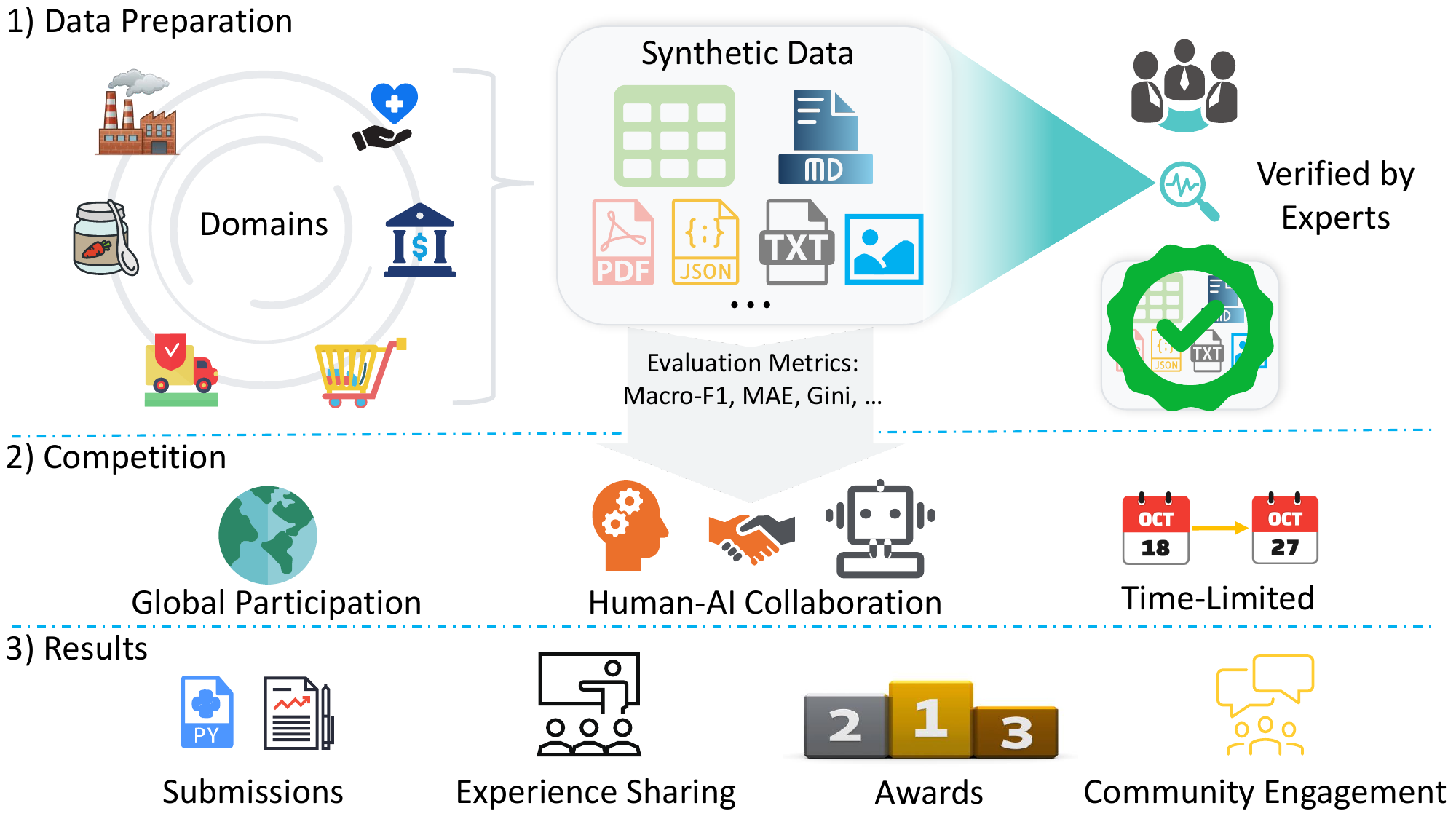} \\
    \caption{Overview of AgentDS.
    The framework comprises three stages: (1) data preparation, where domain-specific challenges are built using realistic synthetic multimodal data and validated by experts; (2) a time-limited competition supporting global participation and human–AI collaboration; and (3) results, including submissions, shared insights, awards, and community engagement, which provide empirical evidence on human, AI, and collaborative performance.
    }
    \label{fig:overview}
\end{figure}

    
	Understanding these differences is important for advancing both AI capabilities and human-AI collaboration.
To address this gap, we introduce 	\textbf{AgentDS}, a benchmark comprising 17 challenges across 6 domains, each grounded in realistic industry problems and built on carefully designed synthetic datasets that reward domain-specific insight. The challenges are constructed so that generic pipelines relying only on off-the-shelf algorithms perform poorly, while approaches that incorporate domain-informed feature engineering and data processing achieve substantially better results. To evaluate these dynamics in practice, we organized a 10-day competition involving 29 teams and 80 participants, enabling a systematic comparison between human–AI collaborative solutions and AI-only baselines. See the overview of AgentDS in Figure~\ref{fig:overview}.

Our inaugural competition reveals three key findings:

\begin{enumerate}
    \item \textbf{Agentic AI systems struggle with domain-specific reasoning.} 
    Current autonomous agents perform poorly on tasks requiring domain-specific insight, particularly when multimodal signals must be incorporated. In practice, several teams that initially experimented with autonomous agent frameworks ultimately abandoned them in favor of interactive human-guided workflows.

    \item \textbf{Human expertise remains essential.} 
    Human data scientists consistently contribute capabilities that AI lack, including diagnosing modeling failures, injecting domain knowledge through feature design and domain-specific rules, and making strategic decisions about model selection and generalization.

    \item \textbf{Human-AI collaboration outperforms either humans or AI alone.} 
    The most successful approaches combine human strategic reasoning with AI-assisted implementation. In these workflows, humans guide the problem-solving process while AI accelerates coding, experimentation, and iteration.
\end{enumerate}

These findings challenge the assumption that advances in agentic AI will soon enable fully autonomous data science. Instead, our results suggest that effective performance on domain-specific tasks continues to rely on human expertise, particularly for problem formulation, domain-specific reasoning, and strategic decision making. AgentDS provides a benchmark for systematically studying these dynamics and highlights the importance of designing systems that support effective 
human–AI collaboration rather than fully autonomous automation. Beyond its role as a benchmark, AgentDS is a replicable framework: the challenge construction pipeline, evaluation protocol, and competition infrastructure can be adapted by other institutions and communities to study human-AI collaboration in their own domain-specific data science settings.

The remainder of the paper is organized as follows. 
Section~\ref{sec:benchmark} introduces the AgentDS benchmark, including its design philosophy, dataset curation process, evaluation framework, the competition setup  and AI baselines.
Section~\ref{sec:findings} presents empirical findings based on both quantitative results and qualitative analysis of participant submissions. 
Section~\ref{sec:limitations} discusses limitations and directions for future work. 
Section~\ref{sec:conclusion} concludes the paper.

	\section{The AgentDS Benchmark and Competition}\label{sec:benchmark}
	
	\subsection{Design Philosophy}
	
	AgentDS is built on three core principles:
	
	\textbf{1. Domain-specific complexity.} We design the challenges so that strong performance requires domain-specific insights. Generic methods yield baseline results at best; competitive performance demands understanding what features matter in each context and what processing steps are appropriate. This design choice deliberately tests whether agents can apply genuine domain reasoning.
	
	\textbf{2. Multimodal integration.}  Real-world data science rarely involves a single tabular dataset. AgentDS therefore provides not only a primary tabular dataset containing the prediction target, but also additional data modalities such as images (e.g., product photos or vehicle condition images), text (e.g., customer reviews or clinical notes), and structured files (e.g., JSON, PDFs, or additional CSV files linked to the main dataset). This design introduces domain-specific complexity that more closely reflects real-world data science challenges.

    	\textbf{3. Real-world plausibility.} While our data is synthesized, the generation process faithfully mirrors genuine relationships found in actual industry data. Each domain's datasets incorporate realistic constraints and correlations that practitioners encounter. We consult the domain literature, including academic papers, industry reports, and practitioner blogs, to ensure that our data reflect authentic patterns and do not contradict established domain knowledge.
	
	\subsection{Benchmark Scope}
	
	AgentDS covers six domains, each selected for its real-world importance, technical challenge, and diversity of required skills. Each domain includes 2-3 challenges spanning classification, regression, and ranking tasks. An overview of the challenges in each domain is presented in Table~\ref{tab:challenges}.  The six domains were selected to span industries where predictive modeling plays a crucial role and where domain knowledge, heterogeneous data modalities, and business-specific evaluation criteria collectively influence modeling strategies. In commerce, demand forecasting and coupon targeting are high-impact problems where behavioral and contextual signals are essential, and product recommendation from visual catalogs benefits substantially from fusing image embeddings with interaction data ~\citep{li2022coupon,liu2023coupon,alamdari2022imagerecsys}. In food production, shelf life estimation requires integrating storage conditions with microbiological growth dynamics, while visual quality control now approaches human inspector accuracy on structured defect detection tasks ~\citep{tarlak2023shelflfe,hemamalini2022foodquality,xiong2024foodcv}. Healthcare challenges center on clinical prediction tasks, such as readmission, emergency department resource consumption, and discharge readiness, where domain-specific feature engineering around comorbidity combinations, vital sign trajectories, and care pathways is decisive ~\citep{iwagami2024readmission,chiu2023ed,pahlevani2024discharge}. Insurance combines structured actuarial data, free-text claims, and image evidence: text-based triage benefits from domain-adapted language models, risk-based pricing demands actuarially sound calibration, and fraud detection must handle severe class imbalance and adversarial adaptation ~\citep{dimri2022claimschanneling,frees2023pricing,aslam2022insurancefraud}. Manufacturing challenges cover predictive maintenance from sensor streams and supply chain delay forecasting, both requiring domain-specific signals~\citep{ayvaz2021predictivemaint,rezki2024scdelay}. Retail banking offers high-volume transaction data where fraud detection and credit default prediction remain challenging due to rare-event class imbalance, and where feature engineering around behavioral proxies requires practitioner expertise~\citep{xu2021loandefault,hashemi2022bankingfraud}.
	
	\begin{table}[htbp]
		\centering
		\caption{An Overview of Challenges in AgentDS Across Six Domains}
		\label{tab:challenges}
		\small
        \resizebox{0.95\textwidth}{!}{%
		\begin{tabular}{@{}ccccc@{}}
			\toprule
			\textbf{Domain} & \textbf{Challenge} & \textbf{Problem} & \textbf{Metric} & \textbf{Additional Modalities} \\
			\midrule
			\multirow{3}{*}{Commerce}
			& Demand Forecasting & Regression & RMSE & Text, CSV\\
			& Product Recommendation & Ranking & NDCG@10 & Image, CSV \\
			& Coupon Redemption & Classification & Macro-F1 & JSON \\
			\midrule
			\multirow{3}{*}{Food Production}
			& Shelf Life Prediction & Regression & MAE & JSON \\
			& Quality Control & Classification & Macro-F1 & Image, JSON \\
			& Demand Forecasting & Regression & RMSE & Text, CSV \\
			\midrule
			\multirow{3}{*}{Healthcare}
			& Readmission Prediction & Classification & Macro-F1 & JSON \\
			& ED Cost Forecasting & Regression & MAE & PDF, CSV \\
			& Discharge Readiness & Classification & Macro-F1 & JSON \\
			\midrule
			\multirow{3}{*}{Insurance}
			& Claims Complexity & Classification & Macro-F1 & Text \\
			& Risk-Based Pricing & Regression & Normalized Gini & Image, CSV \\
			& Fraud Detection & Classification & Macro-F1 & PDF\\
			\midrule
			\multirow{3}{*}{Manufacturing}
			& Predictive Maintenance & Classification & Macro-F1 & CSV, JSON \\
			& Quality Cost Prediction & Regression & Normalized Gini & Image, JSON \\
			& Delay Forecasting & Regression & MSE & JSON \\
			\midrule
			\multirow{2}{*}{Retail Banking}
			& Fraud Detection & Classification & Macro-F1 & JSON \\
			& Credit Default & Classification & Macro-F1 & JSON \\
			\bottomrule
		\end{tabular}
        }
	\end{table}

	\subsection{Data Curation Process}
	
	Creating datasets that are simultaneously realistic, challenging, and informative requires a systematic approach. Our curation pipeline involves four stages as described below. 
	
\textbf{Stage 1: Domain research.} 
For each domain, we identify critical problems where data science provides value, the types of features and data commonly encountered, domain-specific tools and feature engineering practices, and plausible relationships between predictors and outcomes. This research grounds our dataset generation in authentic domain knowledge, ensuring that solving our challenges mirrors solving real industry problems.

	\textbf{Stage 2: Data generation.} We synthesize data using carefully designed data-generating processes that respect the domain constraints identified in Stage~1. Importantly, the generation procedure ensures that strong predictive performance requires domain-specific reasoning rather than purely generic modeling pipelines. To achieve this, we transform certain latent variables that influence the prediction target into additional data modalities (e.g., images), so that effective feature extraction from these modalities requires domain-specific insights. As a result, each challenge dataset consists of a primary tabular dataset containing the prediction target together with additional data modalities that encode complementary information. We iteratively test baseline approaches (e.g., applying XGBoost to the tabular data alone) to verify that they underperform relative to methods that appropriately leverage the additional modalities with domain-specific insights. An example illustrating this process is provided in~~\citep{luo2025agentds}, with a synthetic property insurance dataset where crucial latent variables were embedded in roof images.

	\textbf{Stage 3: Performance bounds and difficulty calibration.} Because we control the data generation process, we can determine the theoretical upper bound on performance by evaluating the score achievable under perfect knowledge of the data-generating mechanism. This allows us to calibrate challenge difficulty and distinguish between fundamental limits and gaps in possible participant approaches. 
	
	\textbf{Stage 4: Documentation and validation.} Each domain includes a \texttt{description.md} file that serves as a comprehensive documentation explaining domain terminology, data sources, and context. We validate that domain experts find the challenges realistic and that the documented information is sufficient (though not prescriptive) for informed approaches. Finally, the data is prepared per domain, meaning that all challenges within the same domain are organized together as a single package.

	\subsection{Evaluation Framework}
AgentDS evaluates submissions primarily based on predictive performance on held-out test data. 
Each challenge is associated with a domain-specific evaluation metric, following those commonly used in practice, as shown in Table~\ref{tab:challenges}.


	\textbf{Quantile scoring.}
	To enable fair comparison across challenges with heterogeneous metrics and scales, AgentDS employs a quantile-based scoring that normalizes performance into a common [0, 1] scale. For each challenge, participants who submit solutions are ranked according to the challenge-specific metric (e.g., Macro-F1, RMSE, normalized Gini coefficient). Let $i$ be the index of a participant who successfully submitted to the challenge, and let $n>1$ denote the number of such participants. The quantile score of participant $i$ is computed as:
	\[
	q_i = \frac{n +1- r_i}{n },
	\]
	where $r_i$ denotes the rank of participant $i$ (with $r_i=1$ indicating the best performance). This transformation ensures that the top performer receives $q_i = 1 $, the worst performer receives $q_i = 1/n>0$, and the intermediate ranks are linearly interpolated. Participants who do not successfully submit to a challenge are scored $0$ for that challenge, ensuring that non-participation always results in the lowest possible score.
	
\textbf{Score aggregation.}
Each domain contains two or three challenges. 
A participant's domain score is the arithmetic mean of their quantile scores across all challenges in that domain. 
The overall score is then defined as the mean of the six domain scores, yielding a single summary measure of cross-domain data science capability. 
This hierarchical aggregation (challenge $\rightarrow$ domain $\rightarrow$ overall) ensures that each challenge contributes equally to the final ranking.

    \textbf{Tie breaking.}
If two participants obtain the same overall score, ties are broken using efficiency indicators: the participant with fewer submissions ranks higher, and if the tie persists, the participant whose final submission occurred earlier ranks higher.
	
    
\subsection{The AgentDS Competition}

The AgentDS competition benchmarks human--AI collaboration performance in domain-specific data science. Participants are allowed to freely use any AI tools, enabling the competition to capture how humans and AI systems interact in realistic data science workflows.

The competition received more than 400 registrations, and participants were allowed to form teams of up to four people. It lasted for 10 days (October 18, 2025 -- October 27, 2025), and a total of 29 teams consisting of 80 participants made successful submissions. During the competition, each team was allowed up to 100 submissions per challenge. After the competition ended, we collected code and reports from participating teams to verify reproducibility and conduct further analysis.

\subsection{AI-Only Baselines}\label{sec:AI-only}

To contrast with the human-AI collaboration performance achieved by competition participants, we evaluate two types of AI-only baselines representing different levels of autonomy: direct prompting baselines using frontier LLMs, and agentic coding baselines using three autonomous coding or data-analysis tools. All AI-only baselines reported in this paper are evaluated after the AgentDS benchmark was finalized and after the public competition had ended. For each baseline, we compute performance using the same evaluation pipeline as human participants. Specifically, the raw metric score obtained by each baseline in each challenge is inserted into the pool of participant scores, and its quantile position is computed as if it had participated in the competition. This produces an interpretable estimate of where each AI-only baseline would rank among human teams.

\subsubsection{Baseline configurations}

\textbf{Direct prompting baselines.}
We evaluate GPT-5.5~\citep{OpenAI2026gpt55}, Claude Opus 4.7~\citep{Anthropic2026opus}, Gemini 3.1 Pro~\citep{GoogleDeepMind2026gemini31}, DeepSeek V4 Pro~\citep{DeepSeekAI2026v4}, and GPT-4o~\citep{OpenAI2024gpt4o} using the same direct prompting protocol. For each challenge, the model is given the official task description together with a lightweight preview of the available data, including tabular files and any additional modalities such as JSON files, images, or PDFs. The model is asked once to produce a complete Python solution for that single challenge to create a valid \texttt{submission.csv}. The generated script is then executed and submitted to the AgentDS evaluation platform to obtain the raw score. For the LLM calls, we use deterministic inference settings whenever supported, and apply only minimal repairs for the generated code when necessary. Full prompt templates and execution details are provided in Appendix~\ref{app:direct-prompting}.

\textbf{Agentic coding baselines.}
We evaluate three agentic coding tools: Claude Code~\citep{Anthropic2025claudecode}, MorphMind~\citep{MorphMind2026}, and Julius AI~\citep{CaesarLabs2023julius}. Claude Code is a command-line coding agent, while MorphMind and Julius AI provide browser-based environments for AI-assisted data analysis and code execution. For consistency, we use Claude Opus 4.7 as the shared base model across the three tools. For each challenge, the tool is given the task description and access to the relevant data, and then autonomously develops and submits a solution. Unlike direct prompting, these tools can iterate during the run by executing code, inspecting intermediate results, and revising their approach for submission. Each challenge is run without human intervention under a 10-minute time budget, and we report the best submitted result obtained within that budget. Implementation details are provided in Appendix~\ref{app:agentic-coding}.

\subsubsection{Performance of AI-only baselines}

\begin{figure}[htbp]
    \centering
    \includegraphics[width=0.92\textwidth]{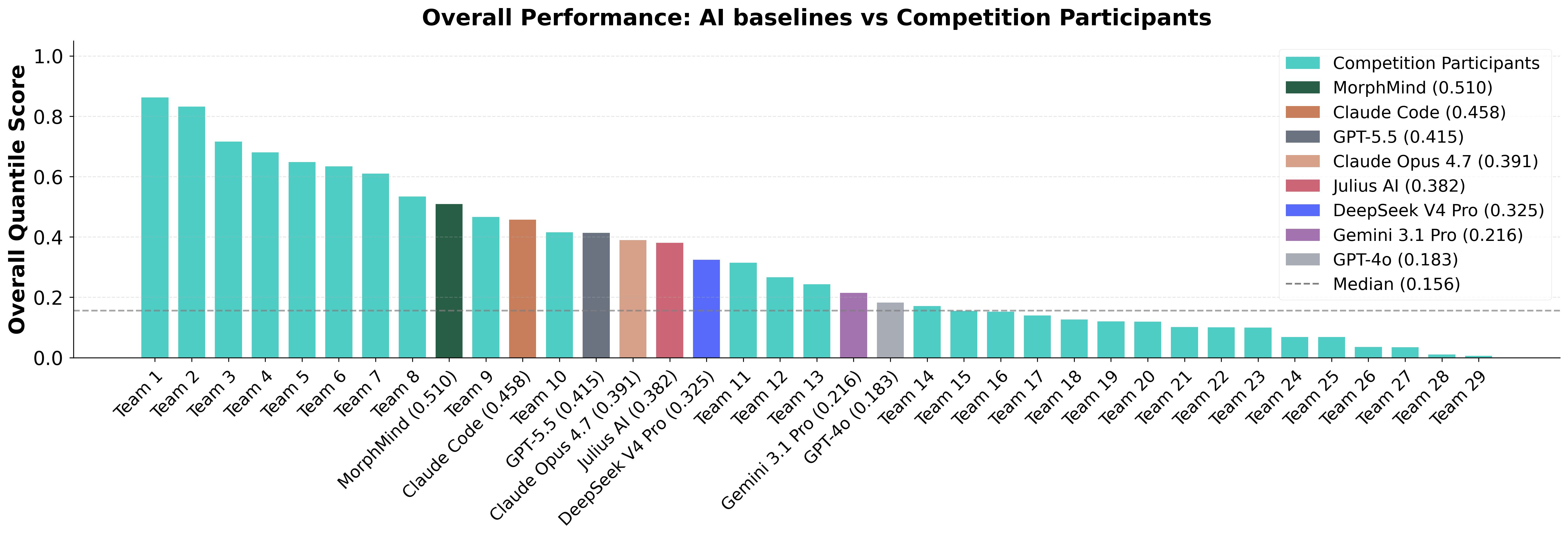}
    \caption{Overall quantile score comparison between all AI baselines and competition teams (n=29). Among agentic baselines, MorphMind (dark green, score: 0.510) ranks 9th, Claude Code (orange-red, score: 0.458) ranks 10th, and Julius AI (rose, score: 0.382) ranks 11th, all above the participant median of 0.156 (dashed line). Among direct prompting LLM baselines, GPT-5.5 (dark gray, score: 0.415) ranks 11th, Claude Opus 4.7 (tan, score: 0.391) ranks 11th, DeepSeek V4 Pro (blue, score: 0.325) ranks 11th, Gemini 3.1 Pro (purple, score: 0.216) ranks 14th, and GPT-4o (light gray, score: 0.183) ranks 14th. Bars are sorted descending by score (Team 1 = best); all AI baselines are inserted at their rank positions. The overall quantile score is the mean of domain scores, each of which is itself the mean of per-challenge quantiles within that domain, with 1.0 indicating best performance and 0.0 indicating non-participation. The result shows that current AI-only baselines, whether using direct prompting or agentic coding, do not match the performance of the top human teams in the competition, highlighting a substantial gap between AI automation and human-AI collaboration.}
    \label{fig:overall_performance}
\end{figure}

Among all baselines evaluated, the three agentic coding baselines obtained overall quantile scores of \textbf{0.510} for MorphMind, \textbf{0.458} for Claude Code, and \textbf{0.382} for Julius AI, corresponding to ranks 9th, 10th, and 11th among 29 teams. Among the direct prompting LLM baselines, GPT-5.5 performs best with an overall quantile score of \textbf{0.415}, ranking \textbf{11th out of 29 teams}, followed by Claude Opus 4.7 (\textbf{0.391}, 11th), DeepSeek V4 Pro (\textbf{0.325}, 11th), and Gemini 3.1 Pro (\textbf{0.216}, 14th). GPT-4o, the weakest direct prompting baseline, achieves an overall quantile score of \textbf{0.183}, ranking \textbf{14th out of 29 teams} and falling near the participant median. Figure~\ref{fig:overall_performance} shows the distribution of overall scores across all participants together with all AI baselines.

\begin{figure}[htbp]
    \centering
    \includegraphics[width=0.9\textwidth]{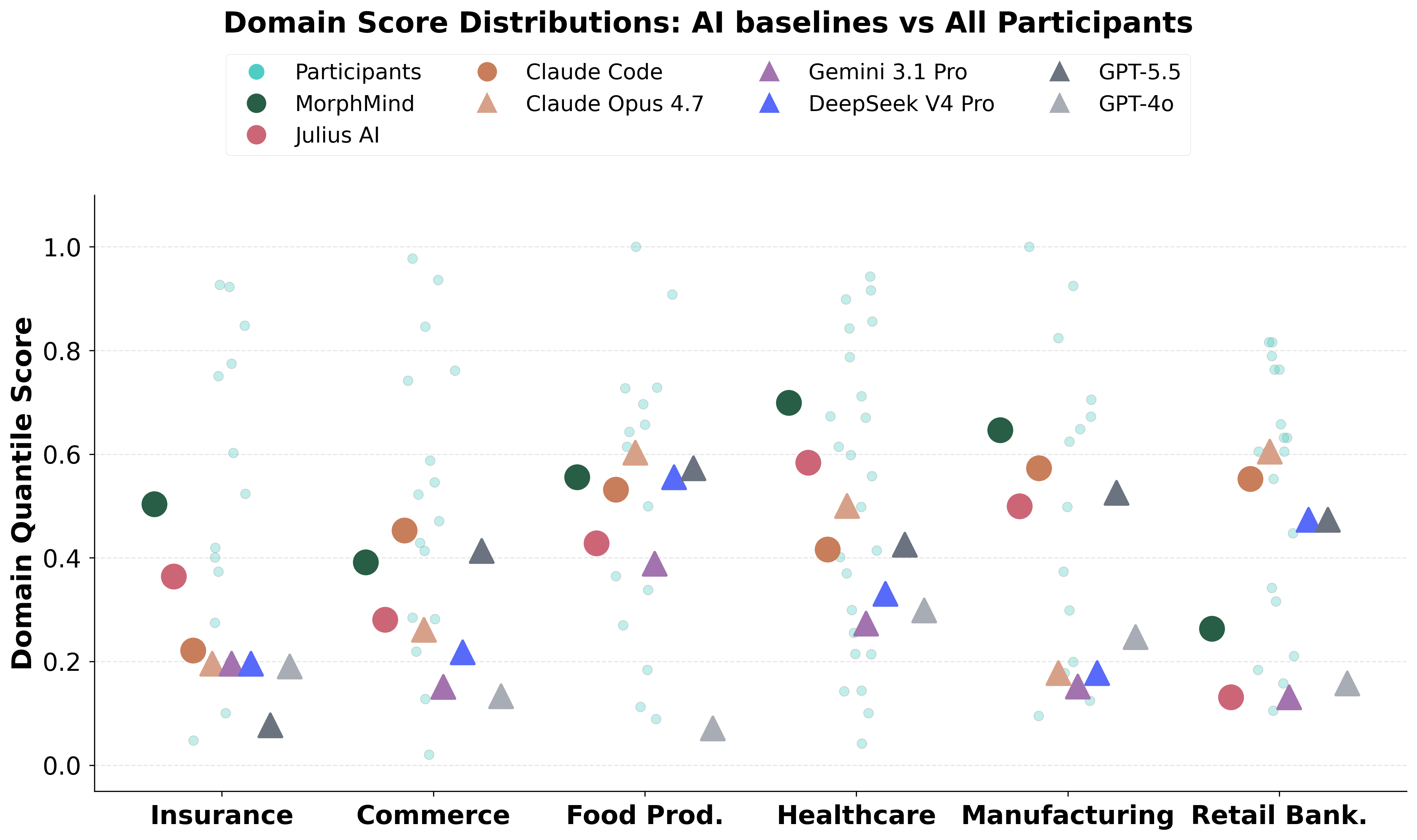}
    \caption{Distribution of domain-level quantile scores across all participants (teal dots), with agentic baselines indicated by filled circles (MorphMind: dark green; Claude Code: orange-red; Julius AI: rose) and direct prompting LLM baselines by filled triangles (Claude Opus 4.7: tan; Gemini 3.1 Pro: purple; DeepSeek V4 Pro: blue; GPT-5.5: dark gray; GPT-4o: light gray). Agentic baselines outperform all direct prompting LLMs across most domains: MorphMind has its highest domain scores in Healthcare (0.699) and Manufacturing (0.647), Claude Code in Retail Banking (0.553), and Julius AI in Healthcare (0.584) and Manufacturing (0.500). Among LLM baselines, GPT-5.5 and Claude Opus 4.7 are competitive in several domains (e.g., Food Production: 0.573 and 0.603 respectively). Nonetheless, all baselines remain well below the top-performing participants in each domain, confirming that current agentic coding tools cannot yet replicate the domain-specific strategies of expert human data scientists.}
    \label{fig:domain_distributions}
\end{figure}

\textbf{Domain-level performance.} Figure~\ref{fig:domain_distributions} illustrates domain-level quantile scores. The GPT-4o baseline performs at or below the domain median across all domains, with particularly weak performance in Food Production (0.071) and Commerce (0.133). Among the stronger direct prompting LLMs, GPT-5.5 achieves competitive scores in Food Production (0.573) and Manufacturing (0.526), while Claude Opus 4.7 performs notably well in Food Production (0.603) and Retail Banking (0.605). DeepSeek V4 Pro and Gemini 3.1 Pro both trail the other frontier LLMs across most domains. The Claude Code agentic baseline substantially improves performance across all domains relative to GPT-4o, achieving its strongest scores in Retail Banking (0.553), Manufacturing (0.573), and Food Production (0.532). Julius AI achieves its highest domain scores in Healthcare (0.584) and Manufacturing (0.500), while performing more weakly in Retail Banking (0.132) and Commerce (0.281). MorphMind achieves the highest domain scores among all baselines in Healthcare (0.699) and Manufacturing (0.647), and is competitive with Claude Code across Food Production (0.556) and Insurance (0.504). Nevertheless, all agentic baselines remain well below the top-performing human teams in every domain.

\begin{figure}[htbp]
    \centering
    \includegraphics[width=0.9\textwidth]{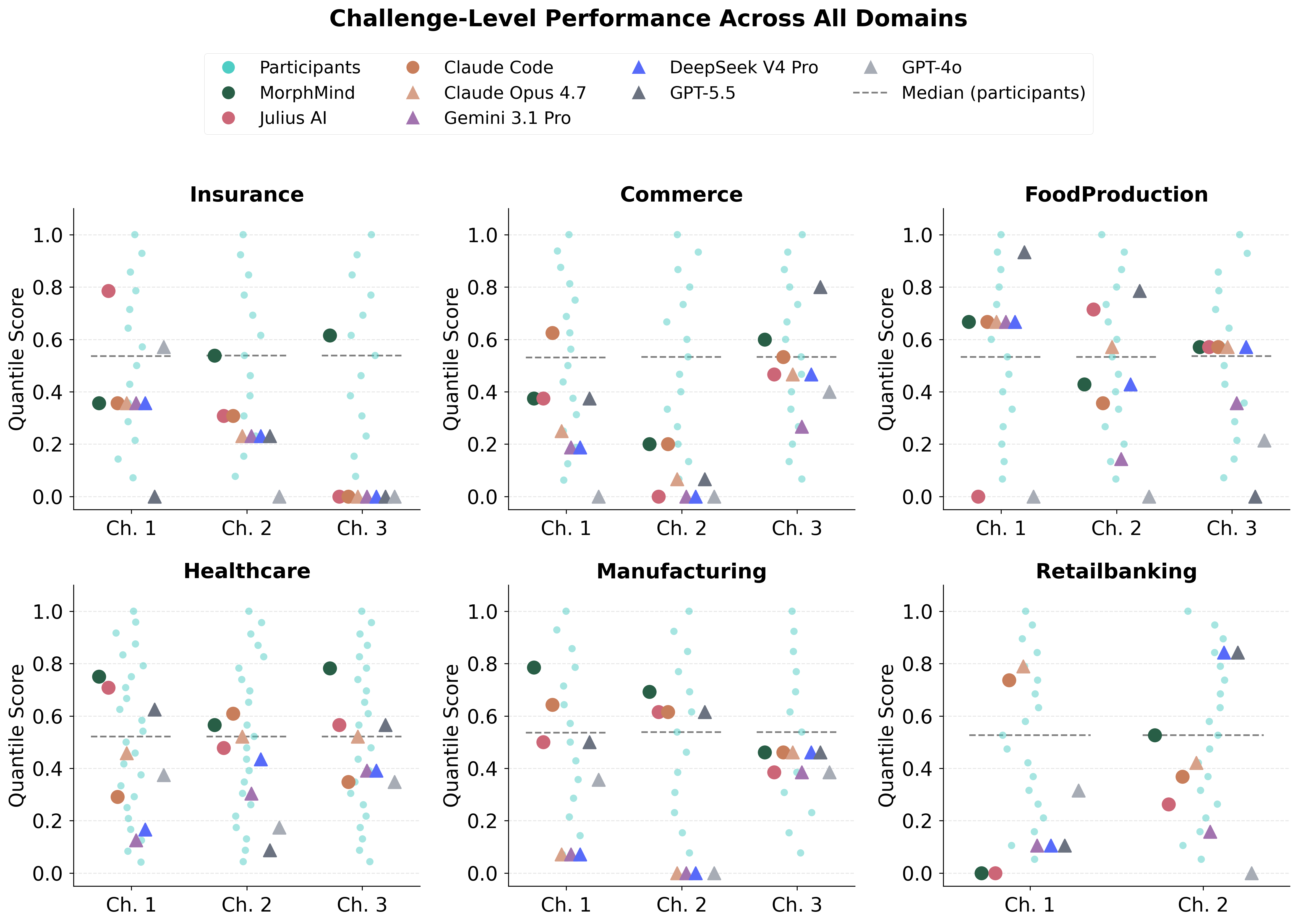}
    \caption{Challenge-specific quantile score distributions across six domains. Teal dots represent participants who submitted for each challenge (zero-score non-participants excluded from display); gray dashed lines indicate per-challenge participant medians among submitters. Agentic baselines are shown as filled circles (MorphMind: dark green; Claude Code: orange-red; Julius AI: rose) and direct prompting LLM baselines as filled triangles (Claude Opus 4.7: tan; Gemini 3.1 Pro: purple; DeepSeek V4 Pro: blue; GPT-5.5: dark gray; GPT-4o: light gray). MorphMind obtains high challenge-level scores on several tasks, including Manufacturing Ch.\ 1 (0.786), Healthcare Ch.\ 1 (0.750), and Healthcare Ch.\ 3 (0.783). Claude Code records its largest gains over GPT-4o in Retail Banking Ch.\ 1 (0.737 vs.\ 0.316) and Commerce Ch.\ 1 (0.625 vs.\ 0.000). Julius AI peaks on Insurance Ch.\ 1 (0.786), Food Production Ch.\ 2 (0.714), and Healthcare Ch.\ 1 (0.708), but scores zero on Food Production Ch.\ 1, Commerce Ch.\ 2, Insurance Ch.\ 3, and Retail Banking Ch.\ 1. Among direct prompting LLMs, GPT-5.5 peaks on Food Production Ch.\ 1 (0.933) and Commerce Ch.\ 3 (0.800), while Claude Opus 4.7 leads in Retail Banking Ch.\ 1 (0.789). No baseline achieves top-quartile performance on every challenge, suggesting that current AI-only approaches still fall short of the best human-AI collaboration teams under this evaluation protocol.}
    \label{fig:challenge_performance}
\end{figure}

\textbf{Challenge-level performance.} Challenge-level results further reveal large performance variability across tasks. As shown in Figure~\ref{fig:challenge_performance}, GPT-4o achieves moderate scores on a small subset of challenges (e.g., Insurance Ch.~1 (0.571), Healthcare Ch.~1 (0.375), Healthcare Ch.~3 (0.348), Commerce Ch.~3 (0.400), Manufacturing Ch.~1 (0.357), and Manufacturing Ch.~3 (0.385)) but obtains near-zero quantile scores on several others. The stronger direct prompting LLMs show uneven profiles: GPT-5.5 scores well on Commerce Ch.~3 (0.800) and Food Production Ch.~1 (0.933) but drops to near zero on several other challenges. Claude Opus 4.7 performs well in Retail Banking Ch.~1 (0.789) and Food Production, but is weak in Manufacturing. Claude Code improves performance on the majority of challenges relative to all direct prompting baselines, particularly in Retail Banking Ch.~1 (0.737) and Commerce Ch.~1 (0.625). Julius AI achieves its highest challenge-level scores in Insurance Ch.~1 (0.786), Food Production Ch.~2 (0.714), and Healthcare Ch.~1 (0.708), but scores near zero on several challenges including Food Production Ch.~1, Commerce Ch.~2, Insurance Ch.~3, and Retail Banking Ch.~1. MorphMind achieves the highest challenge-level scores among all baselines on several tasks, including Healthcare Ch.~1 (0.750), Healthcare Ch.~3 (0.783), and Manufacturing Ch.~1 (0.786), yet no baseline consistently matches the strongest human solutions across all challenges.

Taken together, the baselines demonstrate that while agentic tool use substantially improves AI performance over direct prompting, and that stronger frontier LLMs generally outperform weaker ones in direct prompting settings, all AI-only baselines remain well below the level of the best human data scientists in domain-specific data science. The direct prompting baselines rely on generic modeling pipelines and largely ignore the additional data modalities provided in the challenges. The agentic baselines benefit from iterative experimentation and code execution, but still default to standard modeling strategies and fail to fully exploit the domain-specific signals available in these additional data sources.

These results establish an empirical reference point for interpreting participant outcomes. While the strongest agentic baselines can outperform weaker participants, all AI-only baselines remain below the performance achieved by the strongest teams with human-AI collaboration.

\begin{figure}[htbp]
    \centering
    \includegraphics[width=0.9\textwidth]{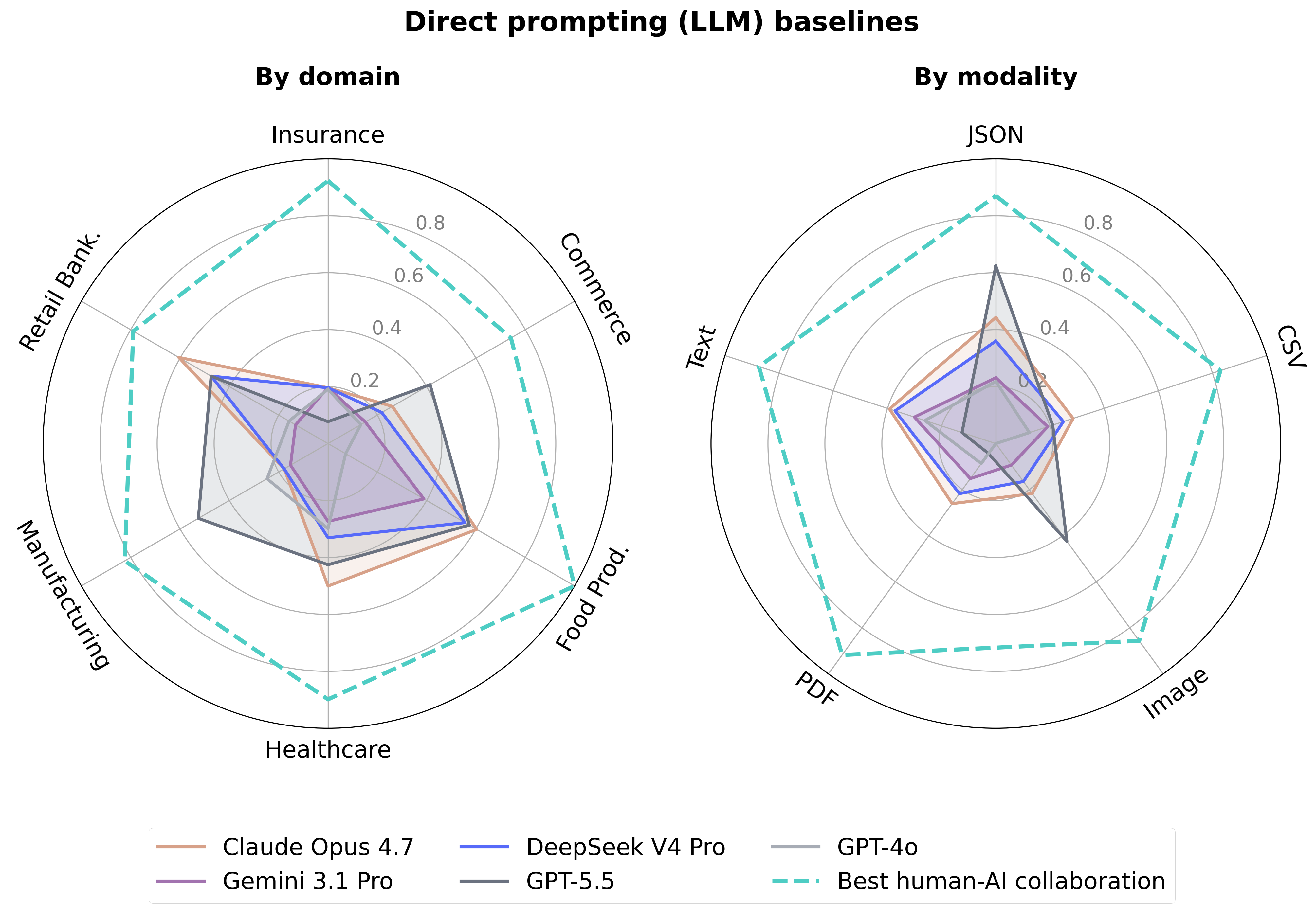}
    \caption{Performance of five direct-prompting LLM baselines (GPT-5.5, Claude Opus 4.7, DeepSeek V4 Pro, Gemini 3.1 Pro, GPT-4o) on the AgentDS benchmark, compared with the best human--AI collaboration team (dashed teal). Each radial axis is a quantile in $[0,1]$: the fraction of non-zero human submissions a baseline outperforms on that challenge, averaged across challenges per domain or modality. \textbf{Left (By domain):} mean per-challenge quantile within each of six domains. \textbf{Right (By modality):} mean per-challenge quantile over all challenges tagged with each additional data modality. GPT-5.5 ranks highest in this cohort (overall 0.415), with its strongest domain results in Food Production (0.573) and Manufacturing (0.526), but near-floor modality scores on PDF (0.040) and Text (0.130). Claude Opus 4.7 (overall 0.391) leads direct-prompting models on Retail Banking (0.605) and performs moderately on JSON (0.440). DeepSeek V4 Pro (overall 0.325) is middling and uneven across domains. Gemini 3.1 Pro (overall 0.216) and GPT-4o (overall 0.183) trail substantially. No direct-prompting model approaches the best human--AI collaboration team (overall 0.860; domain quantiles 0.740--1.000; modality quantiles 0.830--0.920). Structured modalities (JSON, CSV) represent the relative strength of LLMs, occasionally reaching 0.4--0.6, while Image and PDF remain near 0.1--0.3 for most models. LLMs underperform the best human--AI collaboration on every domain and modality, with particularly sharp deficits on multimodal and document-heavy tasks.}
    \label{fig:radar_direct_prompting}
\end{figure}

\begin{figure}[htbp]
    \centering
    \includegraphics[width=0.9\textwidth]{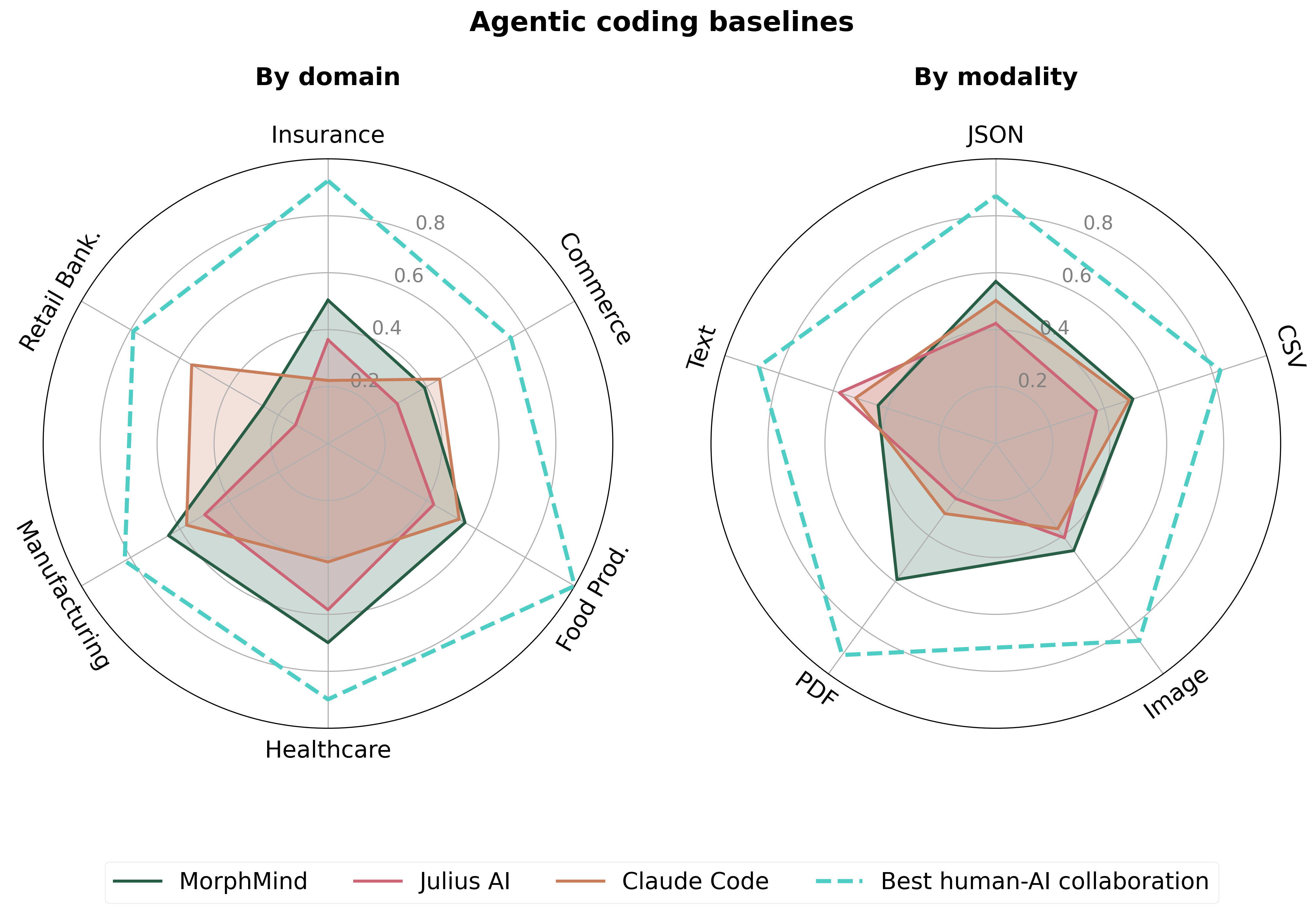}
    \caption{Performance of agentic coding baselines (MorphMind, Julius AI, Claude Code) on the AgentDS benchmark, compared with the best human--AI collaboration team (dashed teal). Each radial axis is a quantile in $[0,1]$: the fraction of non-zero human submissions a baseline outperforms on that challenge, averaged across challenges per domain or modality. \textbf{Left (By domain):} for each of six domains, the plotted value is the mean of per-challenge quantiles within that domain. \textbf{Right (By modality):} for each additional data modality (JSON, CSV, Image, PDF, Text), the value is the mean quantile over all challenges whose additional modality set contains that modality. Among the three agentic coding baselines, MorphMind has an overall mean domain quantile of 0.510, with domain scores of 0.699 in Healthcare and 0.647 in Manufacturing, and modality scores of 0.590 on PDF and 0.570 on JSON. Claude Code has an overall mean domain quantile of 0.458, with higher scores in Retail Banking (0.553) and Commerce (0.450). Julius AI has an overall mean domain quantile of 0.382    , with relatively higher performance on Text (0.580) and lower scores on PDF (0.240) and Retail Banking (0.130). However, all agentic baselines occupy only the inner half of the radar, indicating a large gap between current agentic coding tools and the best human-AI collaboration performance across every domain and modality.}
    \label{fig:radar_agentic}
\end{figure}

\subsubsection{Task-specific data science tools versus general-purpose coding tools}

The agentic coding baselines also allow us to examine whether task-specific data science tools can outperform a more general-purpose coding workflow in AgentDS. Under the same evaluation protocol and using Claude Opus 4.7 as the shared base model, MorphMind obtains the highest overall score among the three agentic baselines, followed by Claude Code and Julius AI. At the same time, Julius AI, although designed for data analysis, does not outperform Claude Code overall. This suggests that task-specificity alone is not sufficient; what matters in AgentDS is whether the workflow supports the kind of iteration needed for domain-specific data science.

The three tools show different strengths under this protocol. Claude Code performs well as a code-centered workflow: it can write scripts, execute them, inspect failures, and improve the scripts. This helps explain why it improves substantially over direct prompting baselines. Julius AI provides a data-analysis-oriented interface and performs relatively well in Healthcare and Manufacturing, but its overall performance is more uneven across domains, with weaker results in Commerce and Retail Banking. MorphMind performs best among the three agentic baselines overall, as its workflow more explicitly separates data exploration, modeling strategy, execution, and result interpretation, which appeared useful for challenges where strong performance required adapting the modeling approach based on intermediate results.

 Taken together, these results suggest that in AgentDS, task-specific data science tools can outperform a general-purpose coding workflow when they support strategy-level iteration in addition to code execution. 

    \section{Empirical Findings from AgentDS}\label{sec:findings}
	
In this section, we present empirical findings based on the quantitative results in Section~\ref{sec:AI-only} and a qualitative analysis of the code produced by the AI-only baselines together with the code and reports submitted by competition participants. Representative examples of competitive submissions are publicly available \href{https://github.com/AnLuo1/AgentDS-Evaluation-Artifacts/tree/master/CompetitiveSubmissions}{\textcolor{blue}{here}}. All evaluation results, including human participant scores, AI-only baseline scores, and modality-level aggregate results, are publicly available \href{https://github.com/AnLuo1/AgentDS-Evaluation-Artifacts/tree/master/EvaluationResults}{\textcolor{blue}{here}}.

    \subsection{AI Agents Struggle with Domain-Specific Reasoning}
	
	Our benchmark reveals concrete evidence of agentic AI limitations. Despite their fluency in code generation and data manipulation, agentic AI consistently underperform on domain-specific data science tasks, as discussed in Section~\ref{sec:AI-only}. Several failure modes emerge:
	
	\textbf{Inability to leverage multimodal signals.} In challenges involving images, such as challenge 2 in insurance and commerce, AI agents fail to extract or appropriately utilize visual features. Human data scientists, by contrast, recognize when image-based signals matter and employ domain-specific computer vision techniques (e.g., DINOv3~~\citep{simeoni2025dinov3}, ResNet50~~\citep{resnet}). This weakness is not limited to image modality: as shown in Figures~\ref{fig:radar_direct_prompting} and~\ref{fig:radar_agentic}, AI baselines are consistently compressed relative to the best human-AI collaboration teams across all additional modalities, including CSV, PDF, text, and JSON, indicating a broad inability to fully exploit the information available in each data type.
	
\textbf{Over-reliance on generic pipelines.} AI tends to default to familiar patterns: loading data, applying standard preprocessing, and training with gradient-boosted models or random forest. While this baseline approach can produce an executable pipeline and works reasonably well for simple tasks, it performs poorly when domain-specific insight is essential, as in AgentDS challenges.

\textbf{Limits of fully autonomous agents.} Fully autonomous agentic approaches remain ineffective for complex domain-specific data science tasks. Several participating teams in AgentDS initially experimented with fully automated agent frameworks but later abandoned them in favor of interactive human-AI collaboration. One team reported that early attempts using autonomous agents with multi-turn tool calls and multi-agent orchestration required extensive prompt engineering and incurred significant API costs, making them difficult to sustain. They ultimately shifted to interactive coding agents, where humans guided the problem solving process while the AI executed coding tasks and explored ideas. This transition improved both practical efficiency and solution quality. Such experiences suggest that current agentic systems are better used as collaborative tools rather than fully autonomous replacements for human data scientists.
	
\subsection{Human Expertise Provides Irreplaceable Value}

Participant reports from the competition reveal a consistent pattern: AI agents accelerated implementation, but the decisions that determined performance were made by humans. The reports highlight four concrete mechanisms through which human expertise contributed value that autonomous agents could not replicate.

\textbf{Strategic problem diagnosis.}
Several top-performing teams explicitly reserved diagnosis for humans and implementation
for AI. Some participants described
a deliberate division of labor in which humans identified the structural weakness of the
current approach, such as miscalibrated peaks, distribution shift between training and test data,
or poorly specified feature interactions, before tasking the AI with implementing the
proposed fix. Others initially pursued fully autonomous
multi-agent frameworks but abandoned them after finding that extensive prompt engineering
yielded diminishing returns. Their eventual approach, interactive human-guided coding
agents, proved substantially more effective. Insights about what worked and what
failed in each domain emerged from human reflection and were then shared back to the agents.

\textbf{Encoding domain knowledge that data cannot reveal.}
Participants frequently constructed features that required domain expertise rather than patterns observable from the data distribution alone. In the healthcare domain, several participants derived features by comparing vital signs against medically defined normal ranges and by engineering indicators capturing stability, volatility, and recovery trends over time. These features reflected clinical protocols that cannot be inferred directly from the data itself. Similar patterns appeared in other domains: some participants incorporated domain-specific business rules, such as credit risk thresholds and inquiry count conditions, which improved model performance beyond what standard machine learning pipelines alone could achieve.

\textbf{Filtering and overriding AI-suggested approaches.}
Multiple teams reported that uncritical acceptance of AI-generated pipelines reduced
rather than improved performance. Some participants observed that AI agents across multiple frontier
models frequently proposed complex feature engineering pipelines that, when
evaluated, lowered their validation scores. They further described a practice of first
reasoning through the problem independently, forming their own hypotheses, and only
then using the agent to implement a human-specified solution. Another team
drew the same conclusion across all seventeen challenges they attempted: domain-driven
feature engineering consistently outperformed blind automation, and no single AI-generated
template generalized across tasks without human adaptation.

\textbf{Human judgment beyond what validation scores reveal.}
Human participants frequently made model-selection decisions that required reasoning beyond simply maximizing validation scores. In several cases, participants deliberately chose models with slightly lower out-of-fold performance because discrepancies between validation and test scores suggested potential overfitting. Such decisions reflect an understanding of generalization risk that cannot be captured by score optimization alone. Participants also exercised caution in how AI tools were used: rather than delegating full control to autonomous agents, many teams conducted experiments manually and used LLMs primarily as assistants for debugging, explanation, or brainstorming. This workflow reflects a broader pattern in which humans retain final judgment in uncertain situations where evaluation metrics alone cannot determine the most reliable modeling strategy.

Taken together, these findings suggest that human expertise contributes more than speed or breadth of search. Humans provide a qualitatively different capability: diagnosing flaws in a model's framing before they appear in the data, injecting domain knowledge that the training distribution does not contain, and maintaining skepticism toward solutions that achieve high validation scores but generalize poorly.

	\subsection{Human-AI Collaboration Outperforms Either Alone}
	
	High-performing approaches in AgentDS competition effectively combine human strategic judgment with AI computational support. This collaboration takes several forms:
	
	\textbf{AI for acceleration, humans for direction.} Successful approaches use AI agents to handle routine tasks, such as data loading, initial exploratory analysis, boilerplate code generation, while humans retain control over strategic decisions: which features to engineer, which models to compare, how to interpret results. This division of labor leverages the strengths of each.
	
\textbf{Iterative human-AI feedback loops.}
Rather than treating AI as fully autonomous, effective collaboration engages tight feedback loops: humans propose approaches, AI implements them rapidly, and humans evaluate results and refine hypotheses. Importantly, these loops are consistently human-initiated. Participants described workflows in which humans judged when results were unsatisfactory, diagnosed the likely cause, and framed the next instruction to the AI. The agent accelerates iteration, but the direction of each cycle is determined by human reasoning.

	\textbf{Complementarity, not replacement.} Human-AI teams excel through complementarity: humans provide domain grounding, causal reasoning, and error correction; AI provides computational power, rapid prototyping, and exhaustive search. Neither alone matches their combined value.
	
	These findings resonate with a growing body of research on human-AI collaboration ~\citep{lai2021towards, inkpen2023advancing, cao2023under, revilla2023human, senoner2024explainable, li2024advanced, fragiadakis2024evaluating}. The central insight is that collaboration quality, meaning how effectively human judgment and AI capabilities are integrated, is as important as the capabilities of either alone. When human-AI collaboration is thoughtfully designed, the resulting partnership can outperform either humans or AI acting alone.

	\section{Limitations and Future Work}\label{sec:limitations}
	
	AgentDS represents an initial step toward rigorous evaluation of AI and human-AI collaboration in domain-specific data science, but several limitations warrant discussion:
	
	\textbf{Synthetic data.} While our data generation process mirrors real-world relationships, it cannot capture the full messiness, ambiguity, and noise of genuine industry datasets. Future iterations may incorporate real (anonymized) datasets where feasible.
	
	\textbf{Limited participation pool.} Our competition drew valuable participation, but larger and more diverse engagement would strengthen findings. We aim to expand outreach in future editions.
	
\textbf{Scope of domains.} Six domains, while diverse, do not exhaust the landscape of applied data science. Future work can expand to additional domains (e.g., energy or other areas of finance) to test the generalization of our findings.

	\textbf{Evolving AI capabilities.} AI systems improve 
  rapidly, and the performance gaps documented here may narrow as agentic 
  systems advance. Nevertheless, the open datasets and evaluation protocol from AgentDS allow future work to re-evaluate new systems on the same challenges, enabling direct comparison over time.

    \textbf{Observational analysis of collaboration.}
Our analysis of human-AI collaboration relies on participant reports, code submissions, 
and qualitative inspection of workflows. While these sources provide rich insight into 
how teams interacted with AI tools, the competition setting does not allow controlled 
experiments on collaboration strategies. Future work could design controlled studies 
that systematically vary the degree of autonomy, prompting strategies, or human oversight 
to quantify which collaboration patterns produce the best outcomes.

\subsection{Replicating AgentDS in Your Own Domain}\label{sec:replication}

Other institutions and communities can reproduce a benchmark and competition like AgentDS by following four components: challenge construction, an evaluation protocol, competition infrastructure, and AI baseline evaluation.

\textbf{1. Challenge construction.} Follow the four-stage curation pipeline 
described in Section~\ref{sec:benchmark}: (i) domain research to identify 
realistic tasks and relevant modalities; (ii) synthetic data generation that 
embeds latent domain signals across modalities; (iii) difficulty calibration 
using known performance bounds; and (iv) documentation of each challenge in a 
\texttt{description.md} file. Crucially, each challenge should be designed so 
that generic off-the-shelf pipelines underperform, requiring domain reasoning 
for competitive results.

\textbf{2. Evaluation protocol.} The quantile-based scoring framework 
(Section~\ref{sec:benchmark}) is metric-agnostic and can be applied to any task for which participant outputs can be evaluated by a comparable performance metric, including but not limited to regression, classification, and ranking tasks across domains.

\textbf{3. Competition infrastructure.}
The competition requires a submission API that accepts \texttt{submission.csv} files for each challenge, validates the submitted predictions, evaluates them against hidden ground-truth test outcomes stored on the server using the official task metric, and returns the resulting score, as well as a leaderboard visible to participants. Both can be hosted using standard institutional infrastructure; our implementation will be made publicly available after acceptance.

\textbf{4. AI baseline evaluation.} Running the direct prompting and agentic 
coding baselines requires only API access to frontier LLMs and, for agentic 
baselines, access to tools such as Claude Code or similar platforms. The 
prompting protocol and evaluation pipeline are fully documented in the Appendix, 
so new baselines can be added without re-running the competition.

\section{Conclusion}\label{sec:conclusion}

AgentDS introduces a benchmark and competition for studying domain-specific data science under realistic conditions. The benchmark comprises 17 challenges across six domains, each designed so that strong performance requires domain knowledge, multimodal reasoning, and thoughtful modeling decisions rather than generic machine learning pipelines. By combining a controlled data generation framework with an open competition setting, AgentDS provides a systematic environment for evaluating both autonomous AI agents and human–AI collaboration for domain-specific data science.

Our results reveal three consistent findings. First, current agentic AI systems struggle with domain-specific reasoning, particularly when multimodal signals and contextual knowledge must be incorporated. Second, human expertise remains essential: participants repeatedly demonstrated the ability to diagnose modeling failures, inject domain knowledge through feature design and domain-specific rules, and make strategic decisions about model generalization. Third, the most successful solutions emerge from human–AI collaboration, where humans guide the problem-solving process while AI accelerates coding, experimentation, and iteration.

These findings suggest that the future of AI in data science may not lie in fully autonomous automation, but in effective human–AI collaboration. Progress therefore depends not only on improving model capabilities, but also on designing AI that better support human reasoning, domain knowledge integration, and iterative problem solving. AgentDS provides a foundation for studying these dynamics and for developing AI that augment, rather than replace, human expertise.

\section*{Disclosure Statement}
AgentDS is financially supported by the Data Science and AI Hub, University of Minnesota, and the Institute for Research in Statistics and its Applications, University of Minnesota. The AgentDS benchmark and competition were developed and organized at the University of Minnesota, with cloud computing services provided through Microsoft Azure. Xun Xian and Jie Ding are affiliated with AIScientists Inc., which developed the MorphMind platform used in the numerical study. 

\section*{Acknowledgments}
	We thank all AgentDS participants for their efforts and insights. 
 	
\bibliographystyle{unsrtnat}
\bibliography{references}
\appendix 
\section{Implementation details for direct prompting LLM baselines}
\label{app:direct-prompting}

This section documents the \emph{direct-prompting} baseline: a single-turn LLM call per AgentDS challenge, followed by automated extraction, execution, and platform submission. The benchmark covers 17 challenges across six domains (Commerce, Retail Banking, Insurance, Healthcare, Manufacturing, Food Production). Each run targets exactly one challenge index within its domain.

\subsection{Chat messages}
\label{app:messages}

Every inference uses two messages:
\begin{itemize}
  \item a fixed \textbf{system} message (\S\ref{app:system-prompt});
  \item a \textbf{user} message assembled from challenge metadata, the full \texttt{description.md}, a data preview, and closing instructions (\S\ref{app:user-prompt}).
\end{itemize}

\subsubsection{System message}
\label{app:system-prompt}

\begin{PromptVerbatim}
You are an expert machine learning engineer.

You are answering **once** with **Python source code only** (no REPL, no tool calls). The benchmark harness will **later** save your code as a `.py` file and execute it in a fresh process.

When that script runs, the harness sets two environment variables (your code must use these -- never hard-code machine-specific absolute paths you might guess):

- `BENCHMARK_DATA_DIR` -- absolute path to the challenge dataset root (all inputs: CSV, JSON, images, PDFs, etc.).
- `BENCHMARK_WORK_DIR` -- absolute path to a writable folder; write **only** `submission.csv` there.

The **task specification** and a **data preview** are already provided in the user message you see now. Your job is to write code that will work when executed later against the real files under `BENCHMARK_DATA_DIR`.

Rules for the generated script:
1. Read inputs with `Path(os.environ["BENCHMARK_DATA_DIR"])` (join relative paths shown in the user message, e.g. `data_dir / "sales_history_train.csv"`).
2. Write `Path(os.environ["BENCHMARK_WORK_DIR"]) / "submission.csv"` using the exact columns and row order required in the user message / description.
3. Prefer `numpy`, `pandas`, `scikit-learn`; other common ML libraries are fine if needed.
4. Do not print or hard-code secrets. Do not call any competition submission API -- the harness submits results separately.

Output format — exactly one program, wrapped like this:

**[CODE]**
```python
# your code
```
**[/CODE]**

Put all imports and logic inside that block. No prose outside the markers.
\end{PromptVerbatim}

\subsubsection{User message}
\label{app:user-prompt}

The user message is assembled from the following template. Braced fields are filled by the harness at run time.

\begin{PromptVerbatim}
## Task

**Challenge:** {challenge.short_name}  
**Index within this domain's challenges:** {challenge.task_number} (1 = first challenge in that domain's `description.md`, etc.)  
**Dataset root (relative to repository root):** `{challenge.data_relpath}/`  
This folder is the same as `BENCHMARK_DATA_DIR` when your generated code runs.

---

## Official specification (`description.md`)

{desc}
{preview_section}
---

## Program you must produce

Follow the system message. The script must run with **no command-line arguments** and create `submission.csv` under `BENCHMARK_WORK_DIR`.

**Single benchmark task:** This run scores **only** challenge index **{challenge.task_number}** in `## Task` above. Do **not** add pipelines or prediction columns for *other* challenge indices from the same `description.md` (e.g. do not bundle task 2 or 3 outputs when this run is task 1). Your final `submission.csv` must match **this** task's required columns and row count.  
**Data prep is different:** Joining, merging, or stacking **input** files (CSVs, JSON, etc.) to build features for that one task is fine and often necessary -- only the **deliverable** must be a single submission for **this** task, not multiple competition tasks at once.

**Structured vitals / labs:** If you parse JSON lists of measurements, drop `null` / missing numeric entries **before** aggregating (e.g. convert to floats and use `np.nanmean` on arrays that may be empty; guard empty lists with a sensible default).
\end{PromptVerbatim}

The field \texttt{\{desc\}} is the full official \texttt{description.md} for the domain. The field \texttt{\{preview\_section\}} is a data-preview block prepared with the following exact header to content generated offline from the dataset directory (directory layout, sampled CSV rows, truncated JSON snippets, and file-name lists for images/PDFs---not inlined):

\begin{PromptVerbatim}
## On-disk data preview (built by the harness from your machine)

The following was read from disk **before** calling the model. It is for context only; at execution time your script must still read from `BENCHMARK_DATA_DIR` (paths below are relative to that directory).

<data preview body>
\end{PromptVerbatim}

The data preview body uses these headings, in order: 
\texttt{\#\# Data layout (on disk today)}; 
\texttt{\#\# Tabular files (CSV)}, which reports column names and the first four rows when present; 
\texttt{\#\# JSON files}, which reports structure and the first two entries when present; and 
\texttt{\#\# Images and PDFs (not inlined)}, which reports up to twelve file names when present.
\subsection{Evaluation protocol}
\label{app:protocol}

Each challenge--model pair is evaluated in four stages.

\textbf{1. Prompt construction.}
Load the official task specification and challenge metadata; attach the on-disk preview; send the system and user messages to the configured provider.

\textbf{2. Obtain code.}
Record the raw model output and extract one Python script and persist it for the run.

\textbf{3. Execute.}
Run the script in a fresh process. We inject \texttt{BENCHMARK\_DATA\_DIR} (dataset root) and \texttt{BENCHMARK\_WORK\_DIR} (writable output directory). The script must read inputs from the former and write \texttt{submission.csv} to the latter. Execution is subject to a configurable timeout (default one hour). Outcomes are classified as: successful submission file present (\texttt{ok}); process failure (\texttt{execution\_error}); clean exit but missing output (\texttt{no\_submission}); or no extractable code (above).

\textbf{4. Submit.}
We authenticates to the AgentDS evaluation platform using the submission credentials (\texttt{AGENTDS\_API\_KEY} and \texttt{AGENTDS\_TEAM\_NAME}) stored in the environment. These are the AgentDS credentials used to upload \texttt{submission.csv} and obtain the raw evaluation score.

\textbf{Inference settings.}
We used the most deterministic configuration supported by each provider. For models whose API accepts a temperature parameter, we set temperature to $0$. For models with optional reasoning or thinking controls, we disabled reasoning/thinking when a no-reasoning or no-thinking mode was available. The exception is Gemini 3.1 Pro, whose API does not provide a no-thinking mode; for this model, we used \texttt{thinking\_level=LOW}, the lowest available thinking setting.

\textbf{Execution repair.}
For generated code that failed during execution, we applied only minimal manual repairs needed to make the submitted solution runnable, such as fixing file paths, syntax errors, column-name references, or submission-file formatting. These repairs were not used to change the model's intended modeling strategy, add new features, or replace the generated solution with a stronger baseline. For GPT-4o on FoodProduction Challenge 2, the output consisted of repeated package imports and contained no task-specific data loading, model fitting, or submission-writing logic; the score for this challenge was therefore reported as 0.

\subsection{Scoring for reported results}
\label{app:scoring}

Raw leaderboard metrics returned by the platform on successful submission are stored per model and challenge. For visualization and aggregate comparisons, each raw score is mapped to a quantile against the distribution of human participants on the same challenge (excluding zero scores), respecting whether higher or lower is better for that task. Domain scores average challenge quantiles within a domain; overall scores average across the six domains.

\section{Implementation details for agentic coding baselines}
\label{app:agentic-coding}

This section documents the evaluation protocol for the agentic coding baselines. We evaluate three agentic coding tools: Claude Code, MorphMind, and Julius AI. Since all three tools support multiple base model choices, we use Claude Opus 4.7 as the shared base model across tools to make the comparison more consistent. All three tools are evaluated in an autonomous setting. After receiving the task prompt, each agent can write and run code, inspect outputs, revise its approach based on intermediate results, and submit predictions to the evaluation platform, without human intervention during the run.

For each domain and challenge index, the agent is instructed to read the official \texttt{description.md}, build a predictive model using the available training data, generate predictions for the corresponding test set, save a valid \texttt{submission.csv}, and submit it through the AgentDS evaluation API. Each run targets a single challenge number within the domain. We record the best submitted score obtained before the agent stops voluntarily or reaches the 10-minute time budget.

The prompt template used for each challenge is shown below. Braced fields are filled in for the corresponding run.

\begin{PromptVerbatim}
You are a data scientist working on a machine learning challenge.

Data directory: {DATA_DIR}
Working directory: {WORK_DIR}

Read description.md at {DATA_DIR}/description.md for full details on columns,
formats, and tasks. Your task is challenge number {N}.

Build the best model you can using the training data, generate predictions for
the test set, save them to {WORK_DIR}/submission.csv, then submit using:

from agentds.client import BenchmarkClient
c = BenchmarkClient('{API_KEY}', '{TEAM_NAME}')
c.authenticate()
result = c.submit_prediction('{DOMAIN}', {N}, '{WORK_DIR}/submission.csv')
print('SUBMISSION_RESULT:', result)

Stop when you are satisfied with your submission.
\end{PromptVerbatim}

\textbf{Claude Code.}
For Claude Code~\citep{Anthropic2025claudecode}, we use the Claude Code CLI, version 2.1.30, with the \texttt{claude-opus-4.7} model in non-interactive autonomous mode. The challenge files are prepared locally before execution. Claude Code is then launched with the prompt above and given a 10-minute time budget for the challenge.

\textbf{MorphMind and Julius AI.}
For MorphMind~\citep{MorphMind2026} and Julius AI~\citep{CaesarLabs2023julius}, we use the same autonomous protocol in their respective cloud-based agentic coding and data analysis environments. For each run, we first instruct the platform to download the relevant AgentDS domain data from Hugging Face, and then provide the challenge-specific prompt using the template above. We record the best submitted result obtained before the run stops or reaches the 10-minute time budget.
\end{document}